# Clarifying Misconceptions in COVID-19 Vaccine Sentiment and Stance Analysis and Their Implications for Vaccine Hesitancy Mitigation: A Systematic Review

Barberia, Lorena[a]; Lombard, Belinda[b]; Norton Trevisan Roman[c], Sousa, Tatiane C. M. Sousa[d]

[a]Department of Political Science, University of São Paulo, Avenida Professor Luciano Gualberto, 315, Room 2067, Cidade Universitária, São Paulo – SP, Brazil Cep 05508-900. E-mail: lorenabarberia@usp.br.

[b]Centre for Systems Modelling and Quantitative Biomedicine, University of Birmingham; Edgbaston B15 2TT, Birmingham, United Kingdom, E-mail: belindalombard@gmail.com Orcid: https://orcid.org/0000-0001-6116-0479.

[c]Escola de Artes, Ciências e Humanidades, University of São Paulo, Rua Arlindo Béttio, 1000, Ermelino Matarazzo, São Paulo – SP, Brazil Cep 03828-000. E-mail: nortontr@gmail.com.

[d]Department of Epidemiology. Institute of Social Medicine, University of Rio de Janeiro State, Rua São Francisco Xavier, 524, Pavilhão João Lyra Filho, 7° andar/blocos D e E, Maracanã, Rio de Janeiro - RJ. CEP 20550-300. E-mail: taticmsousa@gmail.com. Orcid: https://orcid.org/0000-0002-4359-465X.

*Correspondence to:
Lorena Barberia, Department of Political Science, University of São Paulo, Avenida Professor Luciano Gualberto, 315, Room 2067, Cidade Universitária, São Paulo – SP, Brazil Cep 05508-900. E-mail: lorenabarberia@usp.br Phone: +55.11.99499-7664.

**Funding**
LGB [grant number 2021/08772-9] and TCMS [grant number 2022/10997-1] declare financial support from the Sao Paulo Research Foundation (FAPESP). The authors acknowledge the financial support of José Luiz Egydio Setúbal Foundation for underwriting the social media collection, classification, and analysis.

**Abstract**

*Background*

Advances in machine learning (ML) models have increased the capability of researchers to detect vaccine hesitancy in social media using Natural Language Processing (NLP). A considerable volume of research has identified the persistence of COVID-19 vaccine hesitancy in discourse shared on various social media platforms.

*Methods*

Our objective in this study was to conduct a systematic review of research employing sentiment analysis or stance detection to study discourse towards COVID-19 vaccines and vaccination spread on Twitter (officially known as X since 2023). Following registration in the PROSPERO international registry of systematic reviews, we searched papers published from 1 January 2020 to 31 December 2023 that used supervised machine learning to assess COVID-19 vaccine hesitancy through stance detection or sentiment analysis on Twitter. We categorized the studies according to a taxonomy of five dimensions: tweet sample selection approach, self-reported study type, classification typology, annotation codebook definitions, and interpretation of results. We analyzed if studies using stance detection report different hesitancy trends than those using sentiment analysis by examining how COVID-19 vaccine hesitancy is measured, and whether efforts were made to avoid measurement bias.

*Results*

Our review found that measurement bias is widely prevalent in studies employing supervised machine learning to analyze sentiment and stance toward COVID-19 vaccines and vaccination. The reporting errors are sufficiently serious that they hinder the generalisability and interpretation of these studies to understanding whether individual opinions communicate reluctance to vaccinate against SARS-CoV-2.

*Conclusion*

Improving the reporting of NLP methods is crucial to addressing knowledge gaps in vaccine hesitancy discourse.

**Introduction**

Since the onset of the COVID-19 pandemic, there has been considerable effort to understand public opinions towards COVID-19 vaccines and vaccination. Researchers have increasingly used advanced computational tools and approaches, such as Natural Language Processing (NLP), to study discourse on various social media platforms.[1] The heightened attention to studying public health discussions on social media has been motivated by findings from analyses of sentiment and stance suggesting that there is growing polarization surrounding COVID-19 vaccination in discourse on platforms such as Twitter (officially known as X since 2023),[2] Facebook, and Instagram. This polarization reflects disagreements on the urgency, safety, and effectiveness of COVID-19 vaccination and the challenges posed by the COVID-19 "infodemic" - the overwhelming spread of accurate and misleading information. This polarization has hindered public health responses and increased vaccine hesitancy, which contributes to adverse population outcomes, including higher infection and mortality rates.[3]

Researchers have typically employed two key NLP approaches, Sentiment Analysis and Stance Detection, to address the growing polarization in COVID-19 vaccine discourse on social media platforms. Although interrelated, these methods capture distinct yet complementary aspects of public discourse—Sentiment Analysis[4] identifies the overall emotional tone in a message (e.g., positive or negative), while Stance Detection[5] examines the speaker's underlying position (e.g., favorable or unfavorable) toward vaccination. Together, these approaches provide valuable insights into the factors driving vaccine hesitancy and how attitudes toward vaccines are expressed. However, their application has revealed significant methodological challenges, including inconsistent usage and blurred distinctions between sentiment and stance, which can lead to confusion and misinterpretation.[6,7] Precise and accurate conceptualization, categorization, and analysis of these methods are essential, as misclassification or inconsistency reduces their effectiveness in assessments of vaccine confidence.

This study aims to conduct a systematic review of research employing sentiment analysis or stance detection to study the opinions toward COVID-19 vaccines and vaccination spread on Twitter. By critically examining how researchers define, categorize, and codify sentiment and stance, we aim to identify patterns, inconsistencies, and areas for improvement in the application of these methods. We explore whether the objectives stated in these sentiment and stance studies align with their methodologies and evaluate how these approaches contribute to understanding and addressing vaccine hesitancy. By shedding light on the strengths and limitations of sentiment and stance analysis, this work aims to support more effective use of these tools in public health research and communication strategies.

**Data and Methods**

*Study design*

We conducted a systematic review (protocol registered on PROSPERO ID CRD42023457939) to analyze studies investigating COVID-19 vaccine sentiment and stance on Twitter employing supervised machine learning models with Natural Language Processing (NLP). The variables included in our analysis are: (i) the year and journal of publication; (ii) the primary discipline of the publishing journal; (iii) the type of analysis self-reported by the researchers (sentiment vs. stance); (iii) the categories employed for tweet classification; (iv) the definitions provided for the labels or categories used in annotation guidelines; (v) the inclusion or exclusion criteria employed to filter irrelevant tweets and ensure relevance to the research objectives, specifically whether methods went beyond simple keyword filtering to capture contextually relevant data; (vi) whether a category representing neutrality was used and whether it was defined in the study; and, (vii) the central aspect addressed in the results (whether sentiment or stance).

*Search strategy and selection criteria*

The systematic review adhered to the PRISMA guidelines, whose checklist is available in the Supplementary Material. In September 2024, we searched PubMed, Web of Science, and Scopus for papers published from 1 January 2020 to 31 December 2023. For each database, we searched articles published in English using the following search terms and strategy: *((machine learning OR twitter) AND (covid-19 vaccine)) AND ((stance) OR (sentiment))*.

The studies were independently reviewed by two authors (TCMS and BL). For a paper to be included in the review, it must have adhered to all of the following criteria: a machine learning model must have been used to classify tweets exclusively according to either sentiment or stance definitions; the conducted study must have trained the machine learning model (supervised); and the primary focus of the supervised sentiment or stance detection algorithm must have been on COVID-19 vaccines. Discrepancies or disagreements were adjudicated by a third reviewer (LB).

*Data analysis*

We systematically examined each included study to determine how sentiment and stance were conceptualized and analyzed. Specifically, we investigated whether studies using stance detection reported different vaccine hesitancy trends than those using sentiment analysis by assessing how hesitancy was measured and whether efforts were made to avoid potential sources of measurement bias. Sentiment analysis typically studies people's opinions and emotional tone toward an entity. Stance analysis or detection assesses whether the author of a text expresses a position that favors, is against, or has an alternative opinion toward a specific proposition or target. Table 1 outlines the key differences between these study types.

Since sentiment and stance analyses have distinct research objectives, we evaluated whether sentiment analyses used labels consistent with those commonly employed in sentiment research or whether clear annotation definitions aligned with sentiment-related concepts. Similarly, we also assessed if studies self-reported as stance detection employed commonly used stance detection categories or provided definitions explicitly indicating that the annotation identified an opinion toward vaccination or vaccines.

We further investigated whether inconsistencies existed between the self-reported methodological approach in each study and the actual categories used to identify variations in COVID-19 vaccine hesitancy. Specifically, we examined whether studies introduced ambiguity in neutrality as a category, misaligned their findings with their stated method, or made inferences of sentiment based on stance (or vice versa). We also collected the metadata information on each paper for further statistical analysis, including the title, publication year, authors, and venue.

Next, we evaluated the extent to which studies conflated sentiment and stance detection. For example, we assessed whether researchers claiming to analyze sentiment focused on the tone and emotionality of discourse in relation to COVID-19 vaccine hesitancy. Similarly, we examined whether studies describing their analysis as stance detection concentrated on understanding positions or opinions toward COVID-19 vaccines and vaccination, without confusing stance with emotion. However, we observed that in some cases, the two were conflated. For example, studies sometimes inferred conclusions from sentiment that were not aligned with its targeted scope, such as equating negative sentiment with a reluctance to vaccinate. While negative sentiment captures emotional expressions of dissatisfaction or fear, it could also represent feelings of sadness resulting from a lack of vaccines or reduced vaccination coverage, for example. Similarly, inferring emotional tone from stance categories

(e.g., classifying an unfavorable stance as inherently negative or emotional) introduces interpretive biases that can distort study conclusions. To address this, we classified such studies as exhibiting measurement bias.

We identified bias across studies by systematically analyzing classification categories, use of neutrality, potential measurement bias sources, and filtering of irrelevant tweets. The following section details these findings, highlighting how variations in classification choices influenced interpretations of COVID-19 vaccine hesitancy on Twitter.

**Results**

The search in the selected databases retrieved 309 papers: 72 from PubMed, 43 from Scopus and 194 from Web of Science. After 84 duplicates were removed, 225 papers remained for screening. Titles and abstracts were reviewed in the first phase. Based on screening these two categories, 123 studies were excluded. In the second phase, we conducted a full-text review of 102 papers. Following the second-stage review, 51 articles were included in the systematic review, and 51 were excluded, mainly due to being unsupervised studies (n = 25). Figure 1 illustrates the systematic review summary.

The 51 identified papers were published in 36 journals. The highest number of publications occurred in 2022, with 23 papers (41.5% of the sample), followed by 2023, with 18 papers (35.3%), and 10 papers in 2021 (19.6%). No related studies were published in 2020. The journals with more than one paper published are the Journal of Medical Internet Research (n = 8), International Journal of Environmental Research and Public Health (n = 5), Frontiers in Public Health, IEEE Access, International Journal of Advanced Computer Science and Applications and Social Network Analysis and Mining (n = 2). Considering all the identified papers, the primary discipline of the journals where these papers were published was Computer Science (37.2%), followed by Public Health and Medicine (31.3%) or Digital Health (27.5%). Studies published in a Multidisciplinary journal (4%) were the least frequent.

<u>Self-reported study type</u>

Among the 51 studies identified, 39 (76.5%) self-reported as sentiment analysis, while only 12 (23.5%) identified as stance detection studies. Studies on sentiment were most frequent in 2022 (n=21), with stance detection studies comprising only two (8.7%) publications. By contrast, stance studies represented one-third of the total publications in 2023. Regardless of the disciplinary area of the journal, most studies were self-reported as sentiment analyses, with the highest share published in Computer Science (94.7%). There was a relatively higher share of self-reported stance papers in Digital Health (21.4%% %) and Public Health and Medicine (37.5%). Table 2 details these results.

Classification Typology

Table 3 confirms that the categories used to classify tweets varied, particularly in stance detection studies. Sentiment studies primarily employed "positive," "negative," and "neutral" categories, while stance studies included "pro-vaccine," "anti-vaccine," and other nuanced labels. Even though stance detection studies represent a smaller proportion of the research sample (23.5%), there is greater variety in the category labels.

*Mismatch between categories and self-described research type*

The categories to measure COVID-19 vaccine hesitancy should be non-overlapping and mutually exclusive for greater reliability and validity. However, we identified considerable ambiguity in both sentiment and stance classification. Five (12.5%) of the sentiment studies employed stance-like categories. Two of these self-reported sentiment studies[41,42] adopted the classification of tweets annotated as anti-vaccine, neutral, and pro-vaccine, and one study[43] used these same categories and added a fourth class denoted as unrelated. Another sentiment analysis study applied the negative category to identify tweets with and without vaccine rejection causes.[32] Finally, one self-described sentiment study adopted the categories of acceptance, confidence, and perceived barriers to accessing vaccines.[44] In contrast, two stance studies (16.7%) use sentiment analysis-related categories, such as "positive" or "negative."[49,55] Figure 2 highlights these findings by presenting a Venn diagram that illustrates the overlap in categories employed when comparing sentiment and stance studies.

*Missing Definitions and Codebook Availability*

Codebooks that provide definitions are a core element of natural language processing (NLP), most notably in supervised machine learning, where researchers seek to create clear guidelines to identify distinct categories of discourse and evaluate model performance based on an algorithm's precision and accuracy. When coding rules were provided, we were able to verify if categories corresponded to definitions. Only 6 (15%) sentiment analysis studies provided category definitions or codebooks. In contrast, 10 (91%) stance detection studies provided appropriate category definitions. However, we could not access guidelines in most studies (n=35, 68.6%). Thus, in most cases, we could not confirm if the codebook definitions aligned with the category labels. Since studies that lack category definitions may suffer from inconsistent or subjective labeling, there might be a significant bias if codebooks were not built. Of course, an alternative possibility is that they were not cited in the papers. Whatever the reason, a missing description in these studies reduces the research's reproducibility, making it harder to assess the validity of its results.

*Filtering of Irrelevant Tweets*

We investigated whether sample selection bias could be identified due to systematic differences between the sampling frame and the target population in each study. To identify whether this source of bias was present, we examined whether studies reported the inclusion or exclusion criteria employed to filter out irrelevant tweets and ensure relevance to the research objectives, specifically whether methods went beyond simple keyword filtering to capture contextually relevant data. We identified only 11 (28.2%) sentiment studies and 5 (41.7%) stance studies that explicitly described their methods for removing irrelevant tweets after keyword filtering. Furthermore, many studies lacked clear documentation of filtering irrelevant tweets from the chosen sample, potentially introducing noise and reducing classification reliability.

*Measuring neutrality and its association with vaccine hesitancy*

Neutrality is critical in discussions concerning vaccine hesitancy in supervised machine learning. Often, it is argued that a lack of strong opinion can signal a reluctance to vaccinate.[56] Furthermore, neutrality is assumed to capture the discourse of individuals who may not actively oppose vaccination but also do not express clear support; the lack of positions and sentiments in the discourse of these individuals may make them susceptible to being influenced by misinformation. In our analysis, we closely examined how neutrality was defined and used in studies, including whether the annotation guidelines clearly explained neutrality and whether the explanations were related to vaccine hesitancy. Furthermore, we assessed how neutrality was distinguished from other categories, specifically negative sentiment or unfavorable stance. We aimed to identify potential inconsistencies and ambiguities that could impact the interpretation of vaccine-related attitudes.

Neutrality was included as a category in 41 (80.4%) studies. Of this total, 31 (75.6%) were sentiment-related studies.[1,8–12,14–29,31,32,39,40,42,43,57] In these sentiment analyses, neutrality was defined in only four studies, and in all four, it was considered to refer to a lack of explicit attitude toward COVID-19 vaccination.[29,39–41] In the other 10 (24.4%) studies, which account for 83.3% of the stance detection studies,[45,47–50,52,54,55,58,59] neutrality was defined as news-bearing, ambiguous, or lacking strong opinions about COVID-19 vaccination.[47,49] Finally, there were 29 (70.7%) studies that included a neutral category but did not define the neutral category, leaving its interpretation unclear.[8–28,31,32,36,42,43,47,49,57]

*Conflation of Sentiment with Stance in Interpretation*

Even though the objectives of sentiment and stance studies are distinct, researchers often emphasized correlations between sentiment and stance toward COVID-19 vaccines. We identified this more frequently in sentiment studies (n=17, 43.5%). Thus, sentiment studies frequently conflated emotional tone with positional stance. For instance, negative sentiment was often equated with vaccine hesitancy, despite representing diverse emotional expressions, introducing interpretive bias and potential overgeneralization in conclusions. There is also one of the 12 stance detection studies (8.3%) in which the individual's stance towards vaccines was asserted to be associated with sentiment. Figure 3 summarizes these findings. To a lesser extent, although the sample size is smaller, stance detection studies are less likely to associate positions with emotional tone.

To summarize our findings, Table 4 reports the frequency of the five problems that introduced bias in inferences about COVID-19 vaccine hesitancy in Twitter discourse considering the journal's field. Adopting "neutrality" as a classification category and claims about its association with vaccine hesitancy is a frequently occurring practice in sentiment and stance-detection studies (80.4% of the papers). Of the 41 papers where this phenomenon could be found, 12 (29.2%) were published in public health, 11 (26.8%) in digital health, and 16 (39%) in computer science.

The lack of definitions/codebooks was the second most frequent source of error (72.5% of papers). Furthermore, the problem is frequent regardless of the field, as 62.5% of studies published in public health, 71.4% of digital health, and 84.2% of studies in computer science did not include detailed information on definitions or provide codebooks where more detailed annotation guidelines could be retrieved.

The conflation between position and sentiment in interpretation was identified with relatively higher frequency in computer science publications (68.4%). Nevertheless, two (12.5%) studies in Public Health and Medicine and three (21.4%) studies in Digital Health also confused the position of discourse with emotional tone.

The mismatch between employing stance categories to classify emotional tone or emotional tone categories to classify stance was a less common source of bias (only 13.7% of studies). Of the seven studies where this occurred, the highest share of studies appeared in computer science (57%).

**Discussion**

While opinions and emotions towards COVID-19 vaccines and vaccinations are interconnected, various disciplines have repeatedly emphasized that researchers must exercise caution when interpreting one as equivalent to the other.[6,7] Opinions toward vaccines are distinct from emotions toward them, as evidenced by examples such as, *"I am so sad COVID-19 vaccines are not available for my age group yet."* This statement expresses a negative sentiment but conveys a favorable stance. Despite this distinction, we have identified several practices in supervised machine learning studies analyzed in this systematic review that resulted in the conflation of sentiment with a position on vaccine hesitancy. This conflation introduces measurement bias and risks overgeneralizing conclusions, such as if negative sentiment directly corresponds to vaccine hesitancy. Although less frequent, we also identified studies in which positions are conflated with emotional tone, such as studies that imply that unfavorable stances are correlated with negative emotions. Recent NLP research has demonstrated that sentiment and stance often operate independently, particularly in contexts like political and social media discourse.[6, 7]

This bias can be particularly problematic in public health contexts, where effective policies to address vaccine hesitancy depend on a nuanced understanding of its causes. Vaccine hesitancy is not a dichotomous phenomenon but exists on a continuum, as explained by the World Health Organization's (WHO) Strategic Advisory Group of Experts (SAGE).[60] This includes individuals who delay or selectively accept vaccines at specific times, despite availability. SAGE further defines the concept of vaccine hesitancy as delaying acceptance or refusing vaccines despite their availability. Vaccine hesitancy attitudes are influenced by various factors that may affect populations along this continuum differently.[61]

Additionally, SAGE outlined vaccine hesitancy factors as the 3 C's: confidence, which involves trust in vaccine safety, health systems, and policymakers' motives; convenience, which refers to factors like availability, affordability, accessibility, and service quality, influencing vaccine uptake; and complacency, which arises when vaccines are seen as unnecessary due to low perceived disease risks.[60] Considering the 3C approach, sentiment and stance analysis studies on vaccine hesitancy can support vaccine promotion policies by identifying whether unfavorable positions toward vaccination are related to one of these factors. For instance, a lack of trust in pharmaceutical companies (confidence) or the perception that vaccines are unnecessary for children due to their low risk of severe COVID-19 (complacency) are examples of how stance analyses can uncover underlying hesitancy drivers. This occurred, for example, in some populations regarding the perceived low risk of severe COVID-19 in children. Similarly, sentiment analysis can shed light on emotional responses related to hesitancy factors, such as frustration or sadness stemming from limited vaccine availability (convenience), as observed during the early phases of the COVID-19 vaccination rollout. This was observed during COVID-19 vaccination when limited doses were available for all interested individuals. These emotional responses, whether negative (fear, frustration) or positive (hope, joy), provide valuable insights into public sentiment, which can be leveraged to address specific hesitancy triggers.[41] Previous studies, such as Zhao et al.[62], emphasize the role of risk perception in driving emotional responses like fear and anxiety, which subsequently influence coping behaviors during public health crises.

While analyzing sentiment and stance categories, we observed that the "neutral" category was frequently employed but often lacked clear definitions or conceptual consistency. Neutrality was often described as the absence of opinion, but this interpretation is questionable in social media discourse. When individuals choose to express themselves on a topic, it suggests some level of engagement, making true neutrality rare. Instead, many instances labeled as "neutral" may reflect unclear or undefined sentiments or stances rather than an absence of opinion. The emphasis on asserting that discourse is neutral creates bias in vaccine hesitancy research. Most likely, "neutrality" may represent either a lack of engagement or susceptibility to misinformation.

Considering the biases identified in sentiment and stance analysis studies of tweets about COVID-19 vaccination, several recommendations can be outlined to promote research that has greater precision in identifying vaccine hesitancy and reducing measurement error:

> i. a clear exposition of the study's objectives—whether the goal is to identify and analyze sentiments, feelings, opinions, stances, or attitudes toward vaccines;

ii. the availability of comprehensive codebooks that include definitions of the categories used in the study and ensure their conceptual alignment with the study's objectives; and,

iii. explicitly clarifying the conceptual and practical boundaries of the "neutral" category to avoid misclassification and ensure consistent interpretations.

In sum, the results obtained through the systematic review revealed that there has been an explosion in the number of studies dedicated to sentiment and opinion analysis of discourse on COVID-19 vaccination. However, we identified significant biases in these studies, which suggests that their reliability and applicability in shaping understanding of how best to combat vaccine hesitancy is limited. By addressing these issues, sentiment and stance analysis can better support public health efforts, providing actionable insights for strategies to reduce vaccine hesitancy and promote vaccination uptake.

# Figure 1. Flowchart of the Systematic Review

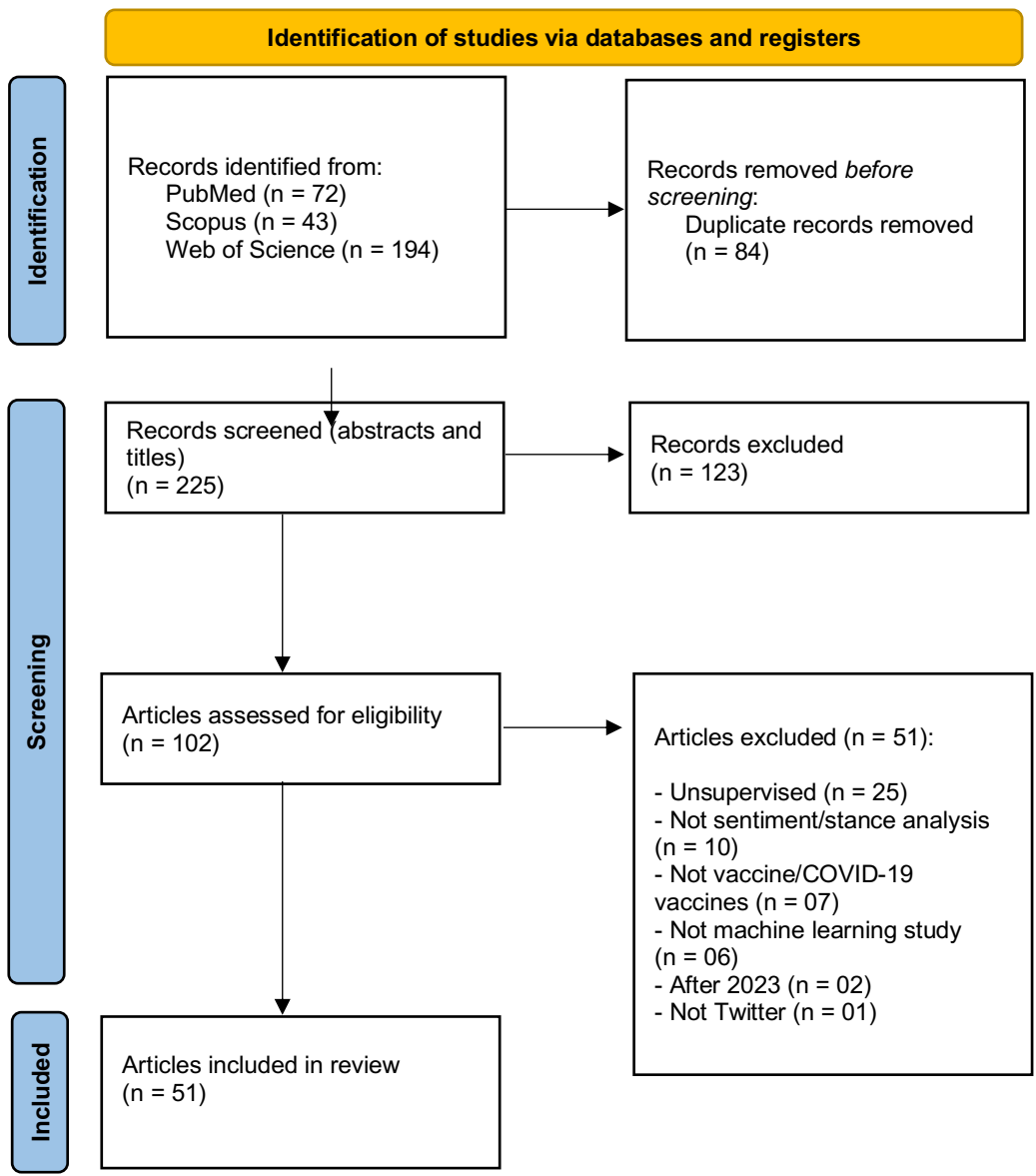

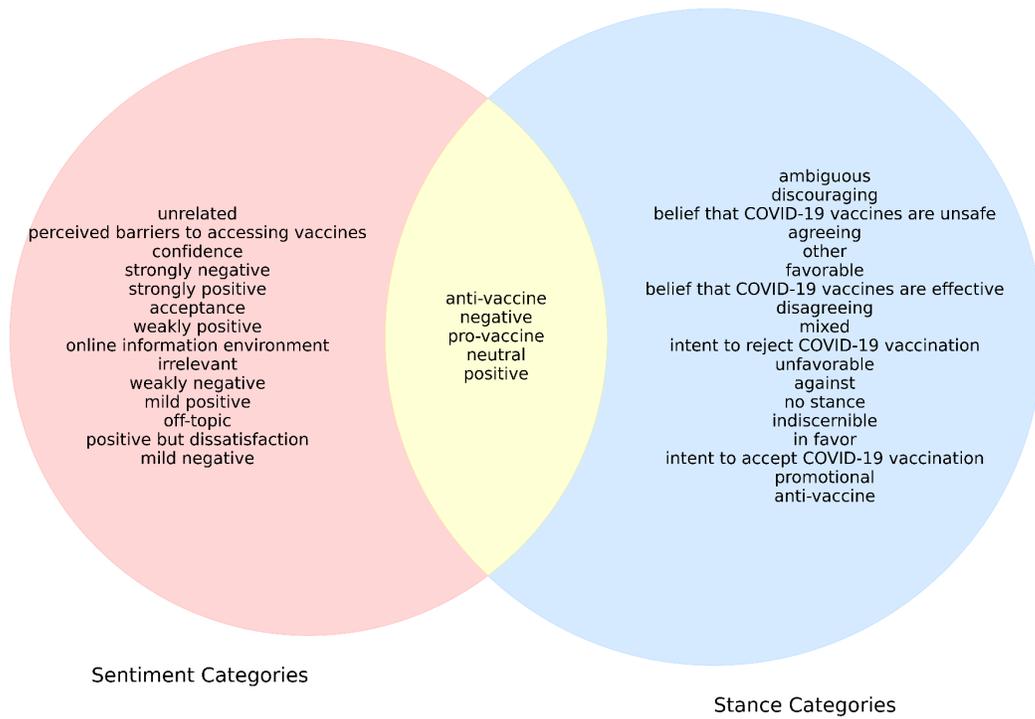

Figure 2. Overlap (yellow) in categories used across self-reported sentiment (pink) and stance studies (blue). Categories such as 'pro-vaccine,' 'anti-vaccine,' and 'neutral' are common to both study types, reflecting ambiguity in classification

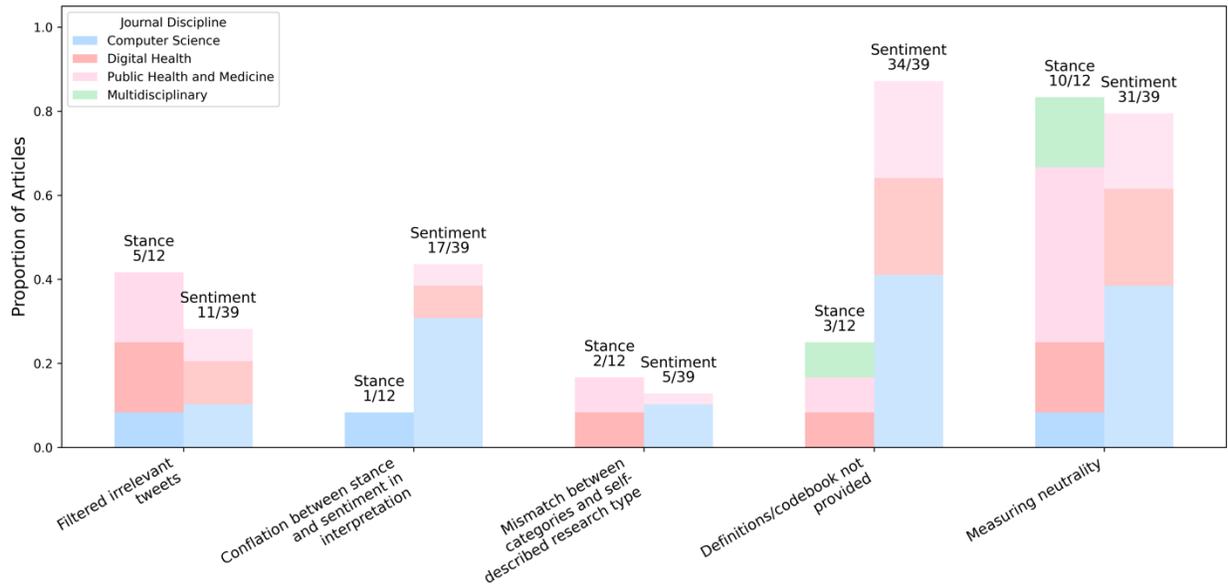

**Figure 3.** Distribution of sentiment and stance studies according to filtering strategies for irrelevant tweets, conflation between stance and sentiment in interpretation, the mismatch between employed categories and self-described study type, provision of definitions/codebooks, and inclusion of neutrality as a category. Values indicate the number of studies relative to the total examined for each classification type. The different colors in the bars represent different journal disciplines.

**Table 1. Categories Commonly Employed in Sentiment Analyses and Stance Detection Studies**

| Study Type | Classification Categories | What Categories Measure |
|---|---|---|
| Sentiment | The discourse was classified using categories that express feelings, such as "positive," "negative," and sometimes "neutral." | Sentiment categories are often described with "positive" reflecting optimistic or supportive emotions (e.g., expressing trust or relief) and "negative" indicating adverse emotions (e.g., fear, anger, sadness). |
| Stance Detection | Discourse was classified using opinion-based categories, such as "pro-vaccine," "anti-vaccine," "neutral," or "conditionally supportive." | For each stance category, opinions are classified considering that "pro-vaccine" indicates endorsement of vaccination, while "anti-vaccine" signifies opposition. |

**Table 2. Distribution of studies according to discipline of the journals and study type**

| Discipline | Journals | Sentiment (n=39) | Stance (n=12) |
|---|---|---|---|
| Computer Science (n=19; 37.2%) | ADCAIJ: Advances in Distributed Computing and Artificial Intelligence Journal (1)[a] | 18 (46.2%)[b] (94.7%)[c] | 1 (8.3%) (5.3%) |
| | Array (1) | | |
| | Computation (1) | | |
| | Computers, Environment and Urban Systems (1) | | |
| | Concurrency and Computation: Practice and Experience (1) | | |
| | Data in Brief (1) | | |
| | Expert Systems with Applications (1) | | |
| | Informatica (1) | | |
| | International Journal of Advanced Computer Science and Applications (2) | | |
| | International Journal of Applied Earth Observations and Geoinformation (1) | | |
| | Journal of Intelligent and Fuzzy Systems (1) | | |
| | Journal of Intelligent Information Systems (1) | | |
| | Mathematics (1) | | |
| | PeerJ Computer Science (1) | | |
| | Social Media + Society (1) | | |
| Digital Health (n=14; 27.5%) | Social Network Analysis and Mining (2)<br><br>The Journal of Supercomputing (1)<br><br>American Medical Informatics Association -AMIA Annual Symposium Proceedings Archive (1)<br><br>BMJ Health & Care Informatics (1)<br><br>Digital Health (1)<br><br>IEEE Journal of Biomedical and Health Informatics (1)<br><br>JMIR Medical Informatics (1)<br><br>JMIR Public Health and Surveillance (1)<br><br>Journal of Medical Internet Research (8) | 11 (28.2%) (78.6%) | 3 (25%) (21.4%) |
| Multidisciplinary (n=2; 4%) | IEEE Access (2) | 0 (0%) (0) | 2 (16.6%) (100%) |
| Public Health and Medicine (n=16; 31.3%) | Bulletin of the World Health Organization (1)<br><br>Disaster Medicine and Public Health Preparedness (1)<br><br>Frontiers in Medicine (1)<br><br>Frontiers in Public Health (2)<br><br>Healthcare (1)<br><br>Healthcare (Basel) (1) | 10 (25.6%) (62.5%) | 6 (50%) (37.5%) |

| | International Journal of Environmental Research and Public Health (5) |
| | Journal of Behavioral Medicine (1) |
| | PLOS ONE (1) |
| | Vaccines (1) |

Notes:

[a] Number of published papers; [b] The percentages in parenthesis refer to the share related to the total in the column. [c] The percentages in parenthesis refer to the share related to the total in the row.

**Table 3. Categories Employed in Self-Reported Sentiment and Stance Detection Studies related to COVID-19 Vaccination**

| Type of Self-Described Analysis | Categories Employed | Publications |
|---|---|---|
| Sentiment (n=39) | Positive, negative, neutral (n=25) | Umair et al., 2023[8]<br>Eom et al., 2022[9]<br>Qorib et al., 2023[1]<br>Mermer et al., 2022[10]<br>Nezhad et al., 2022[11]<br>Agrawal et al., 2022[12]<br>Guo et al., 2023[13]<br>Choi et al., 2022[14]<br>Ogbuokiri et al., 2022[15]<br>Aygun et al., 2022[16]<br>Portelli et al., 2022[17]<br>Ong et al., 2022[18]<br>Kaushal et al., 2023[19]<br>Aslan, 2022[20]<br>Hussain et al., 2022[21]<br>Gbashi et al., 2021[22]<br>Arya et al., 2022[23]<br>Shahzad et al., 2022[24]<br>Alabrah et al., 2022[25]<br>Narasamma & Sreedevi, 2021[26]<br>Jain & Kashyap, 2022[27]<br>Vishwakarma & Chugh, 2023[28]<br>Dupuy-Zini et al., 2023[29]<br>Hussain et al., 2021[30] |
| | Weakly positive, mild positive, strongly positive, weakly negative, mild negative, strongly negative, neutral (n=1) | Umair et al., 2023[31] |
| | Negative (without vaccine rejection causes and tweets with a rejection cause), neutral, positive (n=1) | Alotaibi et al., 2023[32] |
| | Negative, positive (n=1) | Melton et al., 2022[33]<br>Zulfiker et al., 2022[34]<br>White et al., 2023[35]<br>Qorib et al., 2023[36]<br>Roy & Ghosh, 2021[37]<br>Akpatsa et al., 2022[38]<br>Jain et al., 2023[27] |
| | Irrelevant, negative, positive, neutral<br>(n=1) | Ljajic et al., 2022[39] |
| | Negative, off-topic, positive, positive but dissatisfaction, neutral<br>(n=1) | Chen et al., 2022[40] |
| | Anti-vaccine, pro-vaccine, neutral<br>(n=1) | Yin et al., 2022[41]<br>Chen & Crooks, 2022[42] |
| | Anti-vaccine, pro-vaccine, neutral, unrelated to the topic<br>(n=1) | Alhumoud et al., 2023[43] |
| | Acceptance, confidence, online information environment, perceived barriers to accessing vaccines<br>(n=1) | Zhou et al., 2024[44] |
| Stance<br>(n=12) | Against, in favor, neutral<br>(n=4) | Cotfas LA, Delcea C, Gherai R., 2021[45]<br>Cotfas et al., 2023<br>Cotfas et al., 2021[46]<br>Kovacs et al., 2023[47] |
| | Anti-vax, pro-vax, neutral<br>(n=1) | Zaidi et al., 2023[48] |
| | Mixed, negative, positive, neutral<br>(n=1) | Hwang et al., 2022[49] |
| | Favorable, unfavorable, neutral<br>(n=1) | Jing et al., 2021[50] |
| | Intent to accept COVID-19 Vaccination; intent to reject COVID-19 vaccination; belief that COVID-19 vaccines are effective; Belief that COVID-19 vaccines are unsafe<br>(n=1) | Zhou et al., 2023[51] |

| | |
|---|---|
| Promotional, neutral, discouraging, ambiguous, indiscernible (n=1) | Cheatham et al., 2022[52] |
| Agreeing, disagreeing, no stance (n=1) | Weinzieri et al., 2022[53] |
| Anti-vaccine, other (n=1) | To et al., 2021[54] |
| Negative, neutral, positive (n = 1) | Lindelof et al., 2023[55] |

**Table 4. Measurement Bias by Journal Disciplinary Area**

| Discipline | Filtering Irrelevant Tweets | Conflation between stance and sentiment in interpretation | Mismatch between categories and self-described research type | Definitions/codebook not provided | Measuring neutrality |
|---|---|---|---|---|---|
| **Computer Science (n=19)** | 05 (31.3%)[a] (26.3%)[b] | 13 (72.2%)[a] (68.4%)[b] | 4 (57%) (21.0%) | 16 (43.2%) (84.2%) | 16 (39.0%) (84.2%) |
| **Digital Health (n=14)** | 06 (37.5%) (42.9%) | 3 (16.6 %)| (21.4 %) | 1 (14.3%) (7.1%) | 10 (27.0%) (71.4%) | 11 (26.8%) (78.6%) |
| **Multidisciplinary (n=2)** | 0 (0%) (0%) | 0 (0%) (0%) | 0 (0%) (0%) | 1 (2.7%) (50.0%) | 2 (4.9%) (100%) |
| **Public Health and Medicine (n=16)** | 05 (31.3%) (31.3%) | 2 (11.1%) (12.5%) | 2 (28.6%) (12.5%) | 10 (27.0%) (62.5%) | 12 (29.2%) (75%) |
| **Overall (n=51)** | 16 (31.4%)[b] | 18 (35.3%)[b] | 7 (13.7%)[b] | 37 (72.5%)[b] | 41 (80.4%)[b] |

Notes: [a]The percentages in parentheses refer to the share related to the total in the column. [b]The percentages in parentheses refer to the share related to the total in the row.